\documentclass[letterpaper]{article} 
\usepackage{aaai25}  
\usepackage{times}  
\usepackage{helvet}  
\usepackage{courier}  
\usepackage[hyphens]{url}  
\usepackage{graphicx} 
\usepackage{natbib}  
\usepackage{caption} 
\usepackage{algorithm}
\usepackage{algorithmic}

\usepackage{enumitem}
\usepackage{graphicx}
\usepackage{amsmath}
\usepackage{amssymb}
\usepackage{booktabs}
\usepackage{float}
\usepackage{mathtools}
\usepackage{algorithm}
\usepackage{algorithmic}
\usepackage{enumitem}
\usepackage{lipsum}
\usepackage{multirow}
\usepackage{fancyhdr}
\usepackage{newfloat}
\usepackage{listings}
\DeclareCaptionStyle{ruled}{labelfont=normalfont,labelsep=colon,strut=off} 
\floatstyle{ruled}
\newfloat{listing}{tb}{lst}{}
\floatname{listing}{Listing}
\title{MagicNaming: Consistent Identity Generation by Finding a ``Name Space'' \\in T2I Diffusion Models}
\author{
    Jing Zhao\textsuperscript{1}, Heliang Zheng\textsuperscript{3}, Chaoyue Wang\textsuperscript{4}, Long lan\textsuperscript{1}, Wanrong Huang\textsuperscript{1}, Yuhua Tang\textsuperscript{2}\thanks{Corresponding author.}
}
\affiliations{
    \textsuperscript{1}Department of Intelligent Data Science, College of Computer Science,National University of Defense Technology, China\\
    \textsuperscript{2} HPCL,College of Computer Science, National University of Defense Technology, China\\
    \textsuperscript{3} University of Science and Technology of China, China\\
    \textsuperscript{4} University of Sydney, Australian

}

\usepackage{bibentry}

\begin{document}
\twocolumn[{
\renewcommand\twocolumn[1][]{#1}
\maketitle
\begin{center}
    \captionsetup{type=figure}
    \includegraphics[width=\textwidth]{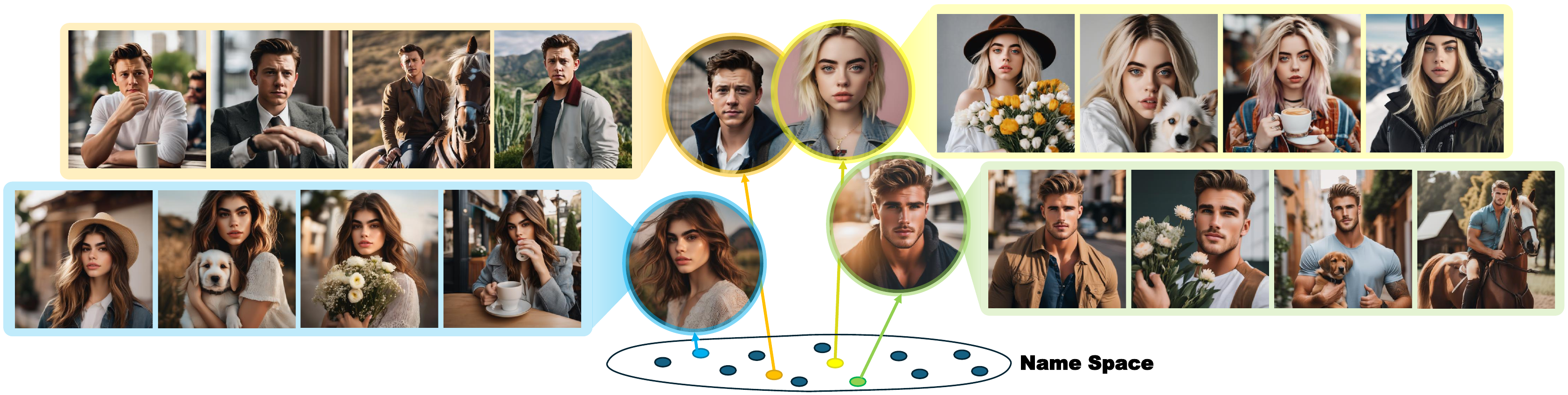}
    \captionof{figure}{Consistent Identity Generation by fetching''names'' from the ''Name Space''.}
\end{center}
}]

\begin{abstract}
Large-scale text-to-image diffusion models, (e.g., DALL-E, SDXL) are capable of generating famous persons by simply referring to their names. \textbf{\textit{Is it possible to make such models generate generic identities as simple as the famous ones, e.g., just use a name?}} In this paper, we explore the existence of a ``Name Space'', where any point in the space corresponds to a specific identity. Fortunately, we find some clues in the feature space spanned by text embedding of celebrities' names. Specifically, we first extract the embeddings of celebrities' names in the Laion5B dataset with the text encoder of diffusion models. Such embeddings are used as supervision to learn an encoder that can predict the name (actually an embedding) of a given face image. We experimentally find that such name embeddings work well in promising the generated image with good identity consistency. Note that like the names of celebrities, our predicted name embeddings are disentangled from the semantics of text inputs, making the original generation capability of text-to-image models well-preserved. Moreover, by simply plugging such name embeddings, all variants (e.g., from Civitai) derived from the same base model (i.e., SDXL) readily become identity-aware text-to-image models. Project homepage: \url{https://magicfusion.github.io/MagicNaming/}.
\end{abstract}

\section{Introduction}
Large-scale text-to-image diffusion models such as DALL-E~\cite{db_51} and Stable Diffusion XL (SDXL)~\cite{stable_diffusion} have revolutionized the field of generative models by enabling the synthesis of images from textual descriptions~\cite{Arar_Gal_Atzmon_Chechik_Cohen-Or_Shamir_Bermano_2023,Chen_Zhang_Wang_Duan_Zhou_Zhu_2023, Gal_Arar_Atzmon_Bermano_Chechik_Cohen-Or_2023, Hinz_Heinrich_Wermter_2022, Li_Wen_Shi_Yang_Huang_2022, Liu_Chilton_2022, Liu_Zhang_Ma_Peng_Liu_2023, Prabhudesai_Goyal_Pathak_Fragkiadaki_2023, Shan_Ding_Passananti_Zheng_Zhao_2023, Xie_Li_Huang_Liu_Zhang_Zheng_Shou_2023}. 
These models have an impressive capability to generate consistent identity images for celebrities simply based on textual name inputs. 
This leads to an intriguing question: Is it possible to extend this capability to generate images of non-known or entirely fictional identities with the same ease, using just a name?

\begin{figure}[t]
  \centering
  \includegraphics[width=\linewidth]{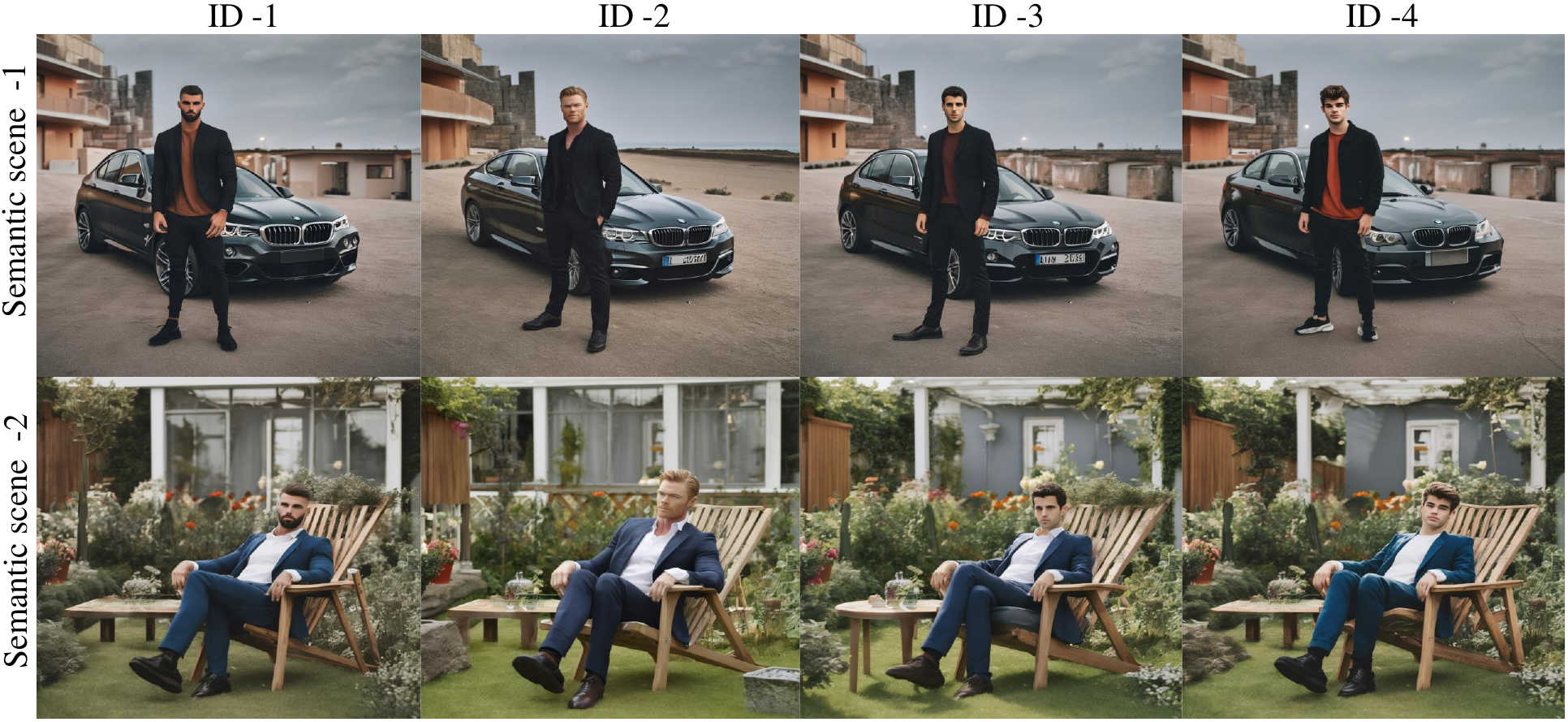}
  \caption{
  Name embeddings and textual semantics are disentangled. 
  }
  \label{fig:decouple}
\end{figure}

In this paper, we investigate the existence of a ``Name Space''($\mathcal N$ Space), aiming to predict a ``name'' for each generic identity and guide the generative model to produce consistent identity generation in a manner that is similar to celebrity name formats. Specifically, we first investigate how celebrity names enable the generative model to produce consistent IDs. Experiments reveal that name embeddings and textual semantics in a text-to-image model are disentangled, e.g., altering name embeddings or their positions within the model does not disrupt the generation of the original semantics. Likewise, maintaining name embeddings while changing semantics does not affect the consistency of the generated IDs. As shown in Figure~\ref{fig:decouple}.

The key to extending the generative model's capability for consistent ID generation from celebrities to generic identities lies in finding corresponding name embeddings for the latter. To this end, we collect a subset from the laion5B dataset that contains celebrity name and image pairs. We conduct text encoding on the names to obtain their respective name embeddings, thereby constructing a large-scale celebrity dataset with their name embeddings. Utilizing this dataset, we train an image encoder capable of predicting the ``name'' for any reference individual. To enhance the alignment with the textual embedding space, our image encoder employs the same CLIP model as used for text embeddings. Building upon this, we refine the output through a trilayer fully connected network to yield the final name embedding prediction. In the inference generation process, a technique analogous to classifier-free guidance was employed, involving the recalibration of name embeddings through the utilization of their mean values. Experiments indicate that our proposed image encoder identifies a $\mathcal N$ Space, wherein each point  corresponds to an individual identity. By embedding this point (i.e., name embedding) into text guidance, we can achieve consistent ID generation for generic identities.

Note that existing research efforts ~\cite{InstantID,facestudio,FastComposer,PhotoMaker,ipadapter} also show the ability to generate consistent identities; however, these approaches predominantly rely on fine-tuning generative models (lora or cross attention layers) on specific datasets, resulting in a bias towards close-up head images. This significantly diminishes the model's capacity, e.g., for scene construction, stylization, emotional editing, and the action control of humans within generated images. In contrast to these approaches, our method does not introduce additional image conditions for text-to-image models, effectively maintaining the original generative capabilities of the diffusion model. Moreover, our approach does not prescribe a fixed generative model; instead, the $\mathcal N$ Space we have identified can be integrated with any variant model based on Stable Diffusion, facilitating the rapid generation of consistent identities. Our contributions can be summarized as follows:

\begin{itemize}[leftmargin=*]

\item We investigated the relationship between prompts and name embeddings, discovering that textual semantics and name embeddings are disentangled. 
Consistent ID generation for generic identities can be achieved by finding its name embedding.
\item We extracted celebrity images from the Laion5B dataset and obtained name embedding by encoding names with text encoder. Through multi-level data filtering and refinement, we constructed the LaionCele dataset, which comprises 42,000 celebrities and approximately 810,000 authentic images.
\item  
We designed and trained an image encoder to accomplish the mapping of any reference portrait to the $\mathcal N$ Space to get the corresponding name embedding. By simply plugging such name embedding into the text embedding, consistent ID generation centered around an specific identity can be achieved. In addition, ID interpolation can also generate fictional characters.

\end{itemize}

\section{Related works}
\label{Relatedworks}

\subsection{Text-to-Image Generation}
Recent years have witnessed a surge of interest in text-to-image generation~\cite{Arar_Gal_Atzmon_Chechik_Cohen-Or_Shamir_Bermano_2023,Chen_Zhang_Wang_Duan_Zhou_Zhu_2023, Gal_Arar_Atzmon_Bermano_Chechik_Cohen-Or_2023, Hinz_Heinrich_Wermter_2022, Li_Wen_Shi_Yang_Huang_2022, Liu_Chilton_2022, Liu_Zhang_Ma_Peng_Liu_2023, Prabhudesai_Goyal_Pathak_Fragkiadaki_2023, Shan_Ding_Passananti_Zheng_Zhao_2023, Xie_Li_Huang_Liu_Zhang_Zheng_Shou_2023,classifier_free} within the field of computer vision, with a plethora of research endeavors propelling rapid advancements in this domain. 
Stable Diffusion~\cite{stable_diffusion} introduced the latent diffusion model, significantly enhancing the generative performance and inference speed of diffusion models for high-resolution images. Building upon this, a myriad of derivative visual tasks have been proposed and explored, including subject-driven text-to-image generation~\cite{gal2022image,dreambooth22,Wei_Zhang_Ji_Bai_Zhang_Zuo_2023,Hua_Liu_Ding_Liu_Wu_He_2023,NeTI,Ku_Li_Zhang_Lu_Fu_Zhuang_Chen_2023,ipadapter}, text-guided image editing~\cite{avrahami2022blended,Hertz_Mokady_Tenenbaum_Aberman_Pritch_Cohen,Tumanyan_Geyer_Bagon_Dekel_2022,Direct_Inversion}, model or subject integration~\cite{LiuLDTT22,Multi_Concept,Cones} and identity-preserved personalization~\cite{facestudio,FastComposer,PhotoMaker}. 
Overall, text-to-image generation is catalyzing a significant wave in the era of AI-generated content.

\begin{figure*}[t]
  \centering
  \includegraphics[width=\linewidth]{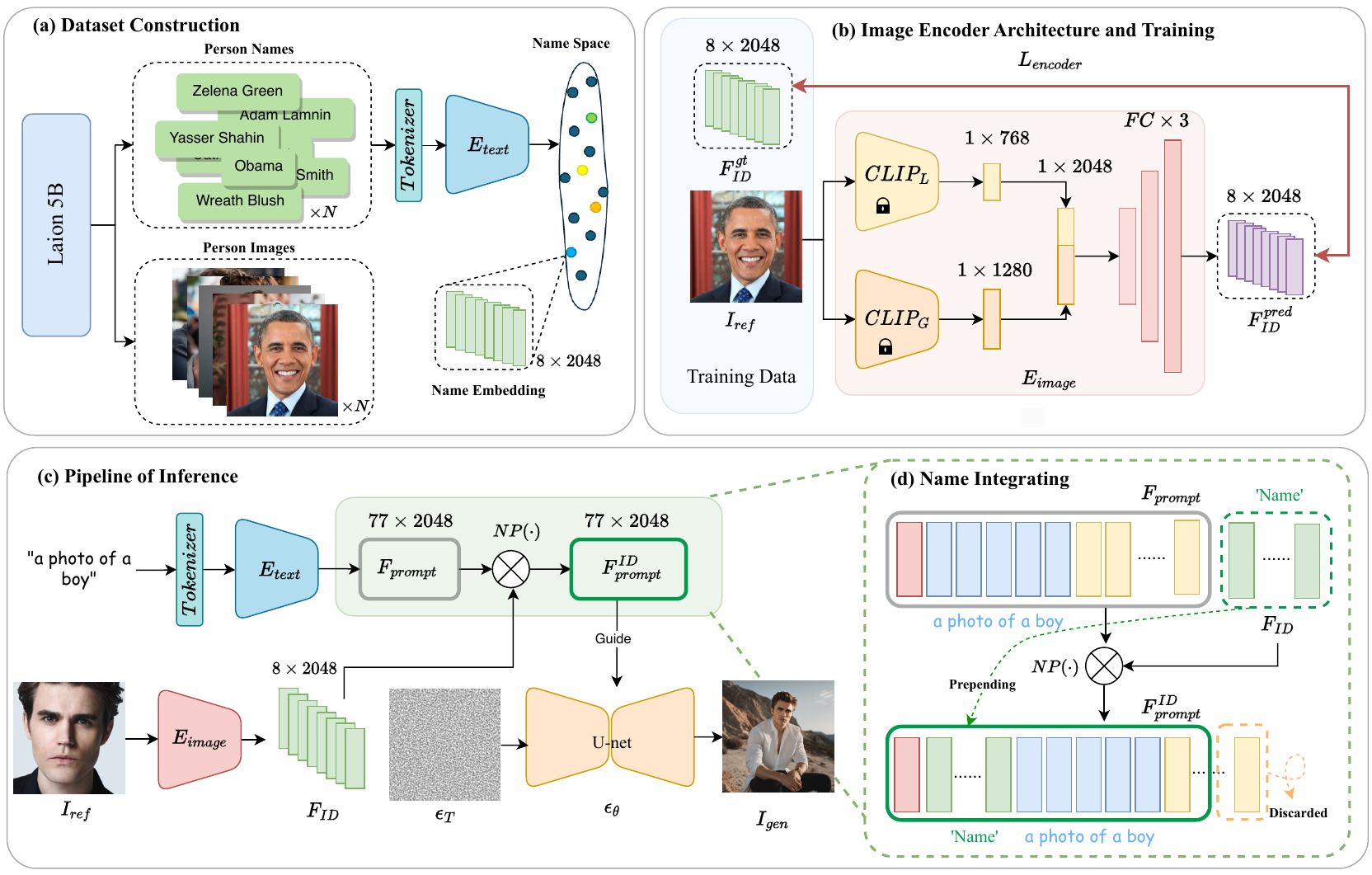}
  \caption{
  Overview of the Proposed Method. 
  (a) Dataset Construction. Celebrity images and their corresponding names were extracted from the Laion5b dataset, with name embeddings generated through a text encoder $E_{text}$ based on the names. 
  (b) Image Encoder Architecture and Training. Features were extracted from input images using two CLIP image encoders, followed by a three-layer fully connected network to produce the ``name'' prediction. Mean Squared Error (MSE) loss was computed against the ground truth name embedding $F_{ID}^{gt}$. 
  (c) Pipeline Inference. The name embedding predicted by the image encoders $E_{image}$ were combined with the original text embeddings by ``name'' Prepending ($NP(\cdot)$) to obtain final embeddings $F_{prompt}^{ID}$. These embeddings were then used to guide a U-net model for denoising. 
  (d) ``Name'' Integrating. 
  There is no need to set a specific placeholder or identify its position, simply inserting the name embedding between the start token(red block) 
  and the first semantic token(blue block) is sufficient to achieve consistent identity generation. 
  The padding tokens(yellow block) exceeding the length of 77 will be discarded. 
  }
  \label{fig:freamworks}
\end{figure*}

\subsection{Identity-preserved Personalization}
The task of Identity-preserved personalization image generation~\cite{facestudio,FastComposer,PhotoMaker,PortraitBooth,InstantID,Zhang_Qi_Zhang_Zhang_Wu_Chen_Chen_Wang_Wen_2022,Pikoulis_Filntisis_Maragos_2023,InstantID} 
demands the generation of personalized images centered around a specific individual, with the consistency of the person's identity being a critical metric of success. 
Facestudio~\cite{facestudio} claims the capability to ``Put Your Face Everywhere in Seconds" highlighting their proficiency in this domain. 
FastComposer~\cite{FastComposer} enables efficient, personalized, multi-subject text-to-image generation without fine-tuning, which uses subject embeddings extracted by an image encoder to augment the generic text conditioning in diffusion models, enabling personalized image generation based on subject images and textual instructions with only forward passes.
IP-Adapter~\cite{ipadapter} is an AI painting tool that generates style-transferred images according to prompts without training Lora, and supports multiple reference images and multiple feature extractions.
PhotoMaker~\cite{PhotoMaker} encodes an arbitrary number of input ID images into a stack ID embedding for preserving ID information. 
PortraitBooth~\cite{PortraitBooth} leverages subject embeddings from a face recognition model for personalized image generation without fine-tuning. It eliminates computational overhead and mitigates identity distortion. 
InstantID~\cite{InstantID} design a novel IdentityNet by imposing strong semantic and weak spatial conditions, integrating facial and landmark images with textual prompts to steer the image generation.
Based on the proposed celebrity foundation, the new identity in INDM~\cite{yuan2023inserting} customized model exhibits better conceptual combination capabilities than previous personalized methods.  However, its image quality and id consistency are not ideal, and it still needs fine-tuning. 
The aforementioned approaches have achieved certain success in consistent identity generation. However, most of these methods destroy the original complex semantic generation ability of the model and rely on fine-tuning.
utilize an image encoder to extract identity information from input images, which is then directly applied to a U-net in a manner akin to text-guided directives. The image generation effect, which combines identity and text guidance, is attained through fine-tuning the U-net. 
employ an image encoder to distill identity features from inputs, subsequently integrating them into a U-net following a protocol analogous to text-guided instructions. The fusion of identity cues and textual guidance for image synthesis is achieved via fine-tuning the U-net. 
The downside of this approach is twofold: on one hand, allowing the U-net to process information beyond text embeddings and updating its parameters can compromise its inherent generative capabilities. On the other hand, fine-tuning the generative model is a one-off process, and the incorporation of identity is not universally applicable across all text-guided diffusion models.
In contrast, our method refrains from adding extraneous learning signals beyond text embeddings,
our method dose not need to fine-tune U-net model, effectively preserving the model's original generative prowess. Moreover, the introduction of $\mathcal N$ Space is compatible with any variant of the SDXL model.

\section{Methods}
\label{method}
\subsection{Decoupling Name Embedding}
It is well-established that diffusion models like SDXL, trained on large-scale image-text datasets, inherently possess the capability to generate images with identity(ID) consistency of celebrities, where the ID information is specified and controlled through the inclusion of names in the prompts.
Our experiments reveal that ID information is localized to the text embeddings associated with name positions in the prompts, i.e., name embedding. As shown in Figure~\ref{fig:decouple}, extracting and integrating these ID-specific embeddings into other semantic embedding allows for the consistency of ID in the generated images.
This phenomenon shows that ID within text embeddings are disentangled from textual semantics, suggesting that the technique of ID consistency through name embedding in image generation is viable.


\subsection{Integrating Name Embedding}

Since the name embedding contains complete personal ID information, it is crucial to devise a method for integrating name embedding into semantic representations in a way that preserves the integrity of the semantic content while ensuring the generated images maintain consistent ID attributes.

``Name'' substitution is an intuitive method that begins by inserting placeholders like ``a man'' or ``a person'' into the original sentence, and then replacing these placeholders with the target ID information in the corresponding name embedding.
However, a limitation of this approach is the inconsistency in placeholder positions across various prompts, which makes identifying the location and length of the placeholder within the text embedding challenging.

In this paper, we employ an alternative approach, namely \textbf{``Name'' Prepending (NP)} Method. As illustrated in Figure. \ref{fig:freamworks}(d),  the target ``name'' $F_{ID}$ is inserted between the start token (the red block) and the first semantic embedding (the blue blocks) of the original text embedding $F_{prompt}$, getting rid of considering the placeholder's position or length. Since the text encoder performs non-semantic padding (the yellow block) for texts with fewer than 77 tokens, 
discarding the excess portion can resolve the embedding length overflow caused by name insertion.
The original text embedding $F_{prompt}=E_{text}(T(p))$,
where $E_{text}$ denotes the text encoder in the diffusion model and $T$ represents the tokenizer which converts a prompt $p$ into tokens. The text embedding resulting from ``name'' prepending is denotes as $F_{prompt}^{ID}$, which can be calculated as follows, 
\begin{equation}\label{id_embedding}
    F_{prompt}^{ID} = NP(F_{ID},F_{prompt})
\end{equation}
where $NP(\cdot)$ represents the operation of ``name'' prepending. 

\subsection{Encoder for Image to $\mathcal N$ Space}
SDXL inherently possesses the capability to generate identity-consistency images of celebrities; the challenge lies in extending this ability to generic identities.
We extract a vast array of celebrity images from the Laion5B dataset and utilize the text encoder to obtain name embeddings corresponding to the names, thereby constructing an extensive identity-name dataset.
Employing this dataset, we trained an image encoder $E_{image}$ to achieve the mapping from individual portraits to a $\mathcal N$ Space, which contains a variety of name embeddings. The architecture of the image encoder is shown in Figure.~\ref{fig:freamworks}(b).

To better approximate the text embedding space, our image encoder incorporates two pre-trained CLIP image encoders $CLIP_L$ and $CLIP_G$, which correspond to the text encoder used in SDXL. For any given reference image $I_{ref}$, features with dimensions 768 and 1280 are extracted using the two CLIP image encoders, respectively. These features are then concatenated and 
subsequently processed through a series of fully connected layers, resulting in a feature with a dimension of 16384. A reshape operation is finally applied to transform the predicted feature into a name embedding $F_{ID}^{pred}$ with a shape of $8 \times 2048$. $F_{ID}^{pred}$ can be denoted as follows,
\begin{equation}\label{eq:f_id}
    F_{ID}^{pred} = E_{image}(I_{ref})).
\end{equation}

The loss function in the training process is defined as follows,
\begin{equation}\label{eq:encoder}
    \mathcal{L}_{encoder} = \mathbb{E}_{\varepsilon}||E_{image}(I_{ref}),F_{ID}^{gt}||^2_2,
\end{equation}
where $F_{ID}^{gt}$ denotes the ground-truth name embedding, $\varepsilon$ denotes the parameters of $E_{image}$.

\subsection{Consistent Identity Generation}

we adopt classifier-free guidance in our sampling process, which extrapolates model predictions toward text guidance and away from null-text guidance. 
Moreover, we design a variant of classifier-free guidance in our name embedding processing by leveraging the mean of name embeddings  $F_{ID}^{mean}$. The specific computation rule is as follows,
\begin{equation}\label{f_mean}
    F_{ID}^{'} = F_{ID}^{mean} + \delta(F_{ID}-F_{ID}^{mean}),
\end{equation}
where $F_{ID}^{mean}$ denotes the average of 42000 name embeddings in the proposed training dataset LaionCele and $\delta$ is used to control the strength of the classifier-free guidance. 
We further introduce a hyper-parameter $\eta$ to adjust the numeric scale of name embeddings, with the specific computation formula as follows,
\begin{equation}\label{f_id2}
    F_{ID}^{''} = \eta \frac{F_{ID}^{'}}{Norm(F_{ID}^{'}))},
\end{equation}
where $Norm(\cdot)$ computes the vector norm of a given tensor in PyTorch. $F_{ID}^{''}$ will be the final name embedding.

To obtain the final generated image, we follow previous work~\cite{ddpm,ddim} to conduct an iterative denoising process. 

Figure.~\ref{fig:freamworks}(c) illustrates the process of consistent identity generation. 
Keeping the predicted name embedding $F_{ID}$ unchanged, variations in the input prompts do not affect the individual identity in the generated images.
A concise summary of the consistent identity generation is presented in Algorithm 1 in the supplementary material .

\section{Experiments}
\label{experiments}
\subsection{Dataset Construction}
In our study, we filtered celebrity images from the Laion5B dataset based on the associated image captions and matched them with corresponding names. Utilizing the prior knowledge of diffusion models, we encoded the names using a text encoder to obtain the respective name embedding. Considering the text encoder's completion mechanism for short texts, we selected the first 8 dimensions of the embedding to serve as the ``name'' for the individuals.
After a rigorous selection and data cleansing process, we retained sample pairs with a minimum of five images per ``name'', resulting in a identity-name dataset comprising 42,000 IDs and 810,000 corresponding images, which we have named LaionCele.
The detailed construction process of the dataset involves numerous procedural intricacies, which are elaborated in the supplementary material.

\begin{figure*}[t]
  \centering
\includegraphics[width=1\linewidth]{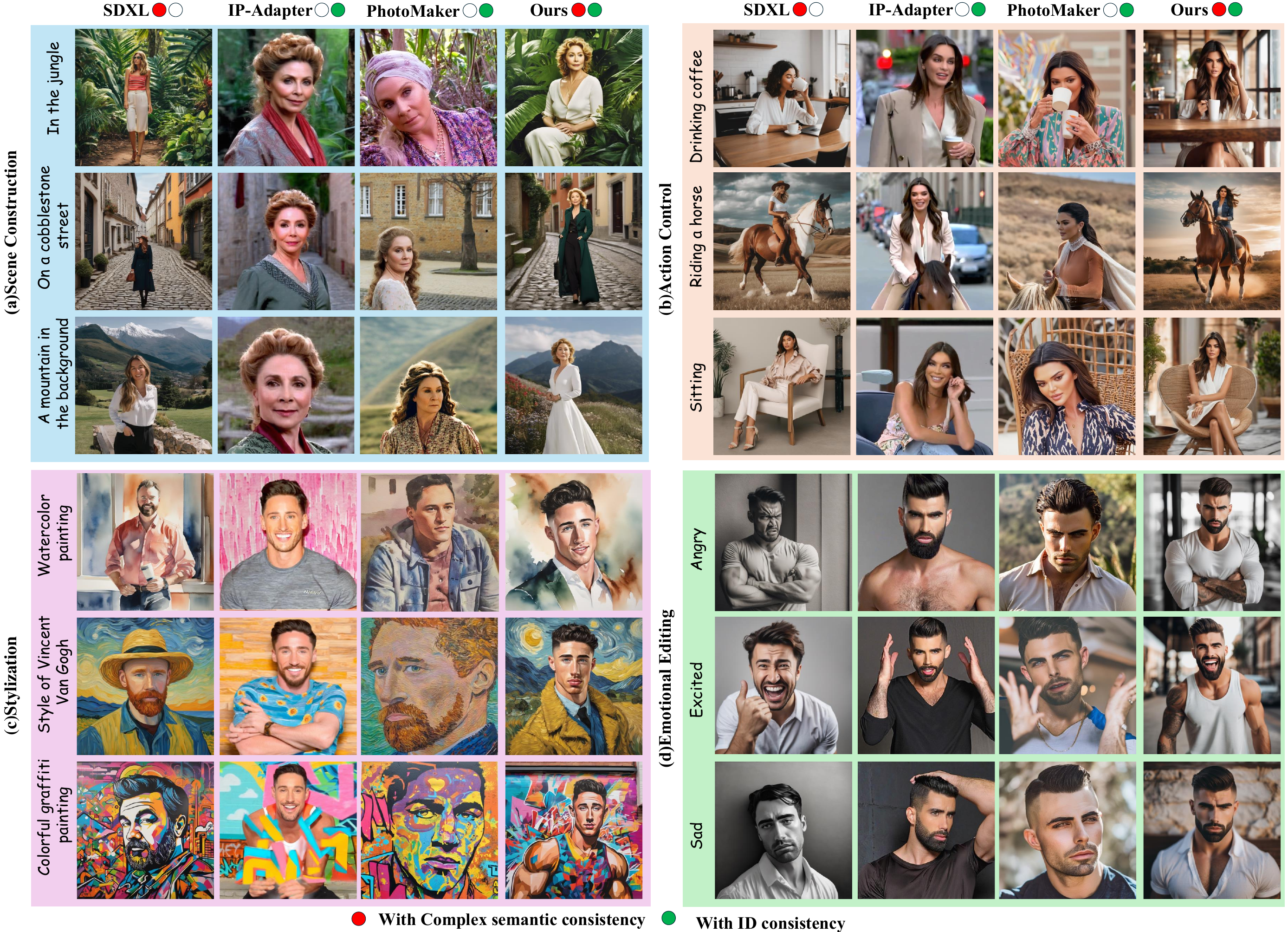}
  \caption{Visualization and Comparison.
   The results concludes four tasks, i.e., scene construction, stylization, action control and emotional editing. 
  Please devote attention to semantic consistency and visual aesthetics of images with the same level of concern as given to ID consistency.
  The results demonstrate that our approach maintains ID consistency while perfectly preserving the original semantic performance (complex semantic consistency) of the generator(SDXL), a feat not achieved by other works.
}
  \label{fig:main-results}
\end{figure*}

\begin{table*}[t]
\caption{The main quantitative evaluation results. Our approach demonstrates superior performance across all metrics.}
\label{tab:main1}
\begin{tabular}{@{}l|ccccc|ccccc|c@{}}
\toprule
\multicolumn{1}{l|}{\multirow{2}{*}{}} & \multicolumn{5}{c|}{CLIP-TI score↑}                  & \multicolumn{5}{c|}{FID score with SDXL↓}                 & \multirow{2}{*}{ID Cons.↑} \\ \cmidrule(lr){2-11}
\multicolumn{1}{l|}{}                  & Style  & Scene & Emotion & Action & All     & Style  & Scene & Emotion & Action  & All     &                           \\ \midrule
SDXL                                   & 40.28 & 40.48  & 27.40 & 37.01  & 36.29  & -      & -      & -       & -        & -       & 0.3365                    \\
IP-Adapter                        & 29.73 & 33.91 & 24.86 & 32.93  & 30.36  & 111.72 & 127.55 & 108.40   & 95.28    & 110.73 & 0.5345                    \\
PhotoMaker                          & 40.37 & 40.72  & 29.11 & 36.93  & 36.78 & 76.11  & 106.68 & 109.58  & 88.27    & 95.16  & 0.5466                    \\ \hline
Ours                                   & \textbf{40.57} & \textbf{43.45} & \textbf{29.31}  & \textbf{39.63}  & \textbf{38.24} & \textbf{49.52}  & \textbf{57.98}  & \textbf{73.25}   & \textbf{46.25}    & \textbf{56.75}   & \textbf{0.5624}                    \\ \bottomrule
\end{tabular}
\end{table*}

\subsection{Results and Comparison}

\subsubsection{\textbf{Evaluation Setup}}
To demonstrate the effectiveness of our method, we conducted both quantitative and qualitative assessments, comparing and analyzing them against relevant works. e.g., IP-Adapter~\cite{ipadapter} and PhotoMaker~\cite{PhotoMaker}.  For the \textbf{testing prompt}, we classified them into four tasks: (a)scene construction, (a)action control, (c)stylization and (d)emotional editing. These prompts present greater challenges compared to simple directives such as changing clothes or wearing hats.
It is noteworthy that (a)scene construction and (b)action control pursue the generation of images that contain complete human figures and rich scene semantics rather than merely headshots. For the \textbf{evaluation images}, we meticulously selected person images that SDXL could not generate with consistent ID based solely on name text, thereby mitigating the influence of the generative model's inherent person recognition capabilities on ID consistency of generated images.
Ultimately, for each evaluation task, we produced 1,000 test images for each method. The detailed configuration of the \textbf{evaluation dataset} (prompts and images) and the specific \textbf{implementation details} of the methodology are elaborated in the supplementary material.

\subsubsection{\textbf{Visualization and Comparison}}

The visual outcomes of the aforementioned tests are partially illustrated in Figure.~\ref{fig:main-results}, with additional results presented in the supplementary material.
SDXL~\cite{stable_diffusion}, serving as the baseline model, demonstrates its inherent complex semantic generative capabilities for the respective tasks. Given that SDXL lacks identity insertion functionality, we facilitate person image generation by inserting the corresponding personal names into the prompt.
The outputs indicate that SDXL inherently lacks the capability to generate consistent IDs for generic identities.
Subsequent sections will provide a comparative analysis for each task individually.

\textit{\textbf{(a)Scene Construction.}}
Applications of scene construction with close-up head are highly constrained.  
Effective scene content and character portrayal are crucial for applications like print advertising and film production.
As shown in Figure.~\ref{fig:main-results}(a), both IP-Adapter and PhotoMaker primarily generate portraits from the chest or neck up, with scene content corresponding to the text being incorporated only by adding relevant elements to the head's background. 
These approaches fall short in comprehensive scene generation.
Conversely, our approach consistently produces complete character representations and well-composed scenes in various contexts, with high-resolution images, semantically rich environments and visual appeal, demonstrating substantial creative potential in role-based scene construction. 

\textit{\textbf{(b)Action Control.}}
The inherent capability of the SDXL to generate human action scenes is notably aesthetically pleasing and impressive, e.g., ``a woman is riding a horse''. However, after fine-tuning, most models show a significant drop in this ability. As observed in Figure.~\ref{fig:main-results}(b), the generated images by PhotoMaker and IP-Adapter 
remain constrained to upper-body of individuals.
resulting in incomplete scenes lacking in visual appeal. In contrast, our approach not only maintains identity consistency but also delivers stunning visual effects, with complete human actions and a semantically rich scene representation.

\textit{\textbf{(c)Stylization.}}
One of SDXL's key strength lies in its ability to synthesize diverse-styled imagery from textual descriptions alone.
We conducted tests across several classic styles, with the results displayed in Figure.~\ref{fig:main-results}(c). 
The results show that the IP-adapter has negligible stylization capability. PhotoMaker, while integrating stylistic features, diverges from intended styles and is plagued by poor image fidelity and content ambiguity.
Even more, instances of failed identity consistency have been observed in PhotoMaker.
Conversely, our method maintains SDXL's inherent stylization proficiency, producing stylized images with superior quality, clear content and consistent identity.

\textit{\textbf{(d)Emotional Editing.}}
Our method maintain the intrinsic generative power of the generator model (e.g., SDXL), which enables us to leverage sophisticated semantic manipulation features, including the ability to edit emotional expressions.
As illustrated in Figure~\ref{fig:main-results}(d), our method successfully facilitates the generation of diverse emotional states, while ensuring the synthesized identities remain consistency.
In contrast, the facial expressions of characters generated by IP-Adapter and PhotoMaker exhibit less pronounced emotional variations.

\begin{figure}[t]
  \centering
    \includegraphics[width=\linewidth]{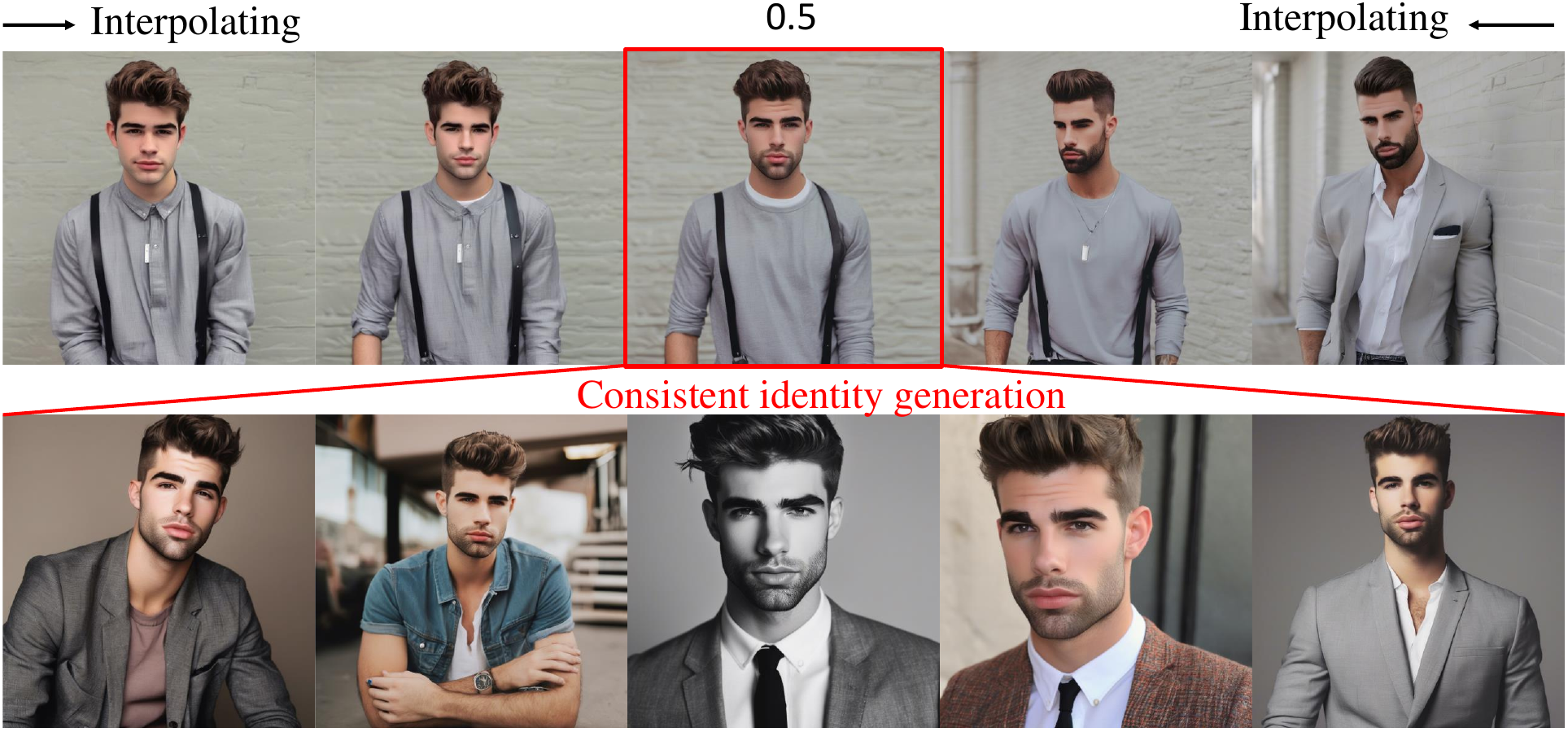}
  \caption{A fictional character is created by interpolating, and the fictional character supports consistent identity generation.
  }
  \label{fig:chazhi}
\end{figure}

\begin{figure}[t]
  \centering
   \includegraphics[width=\linewidth]{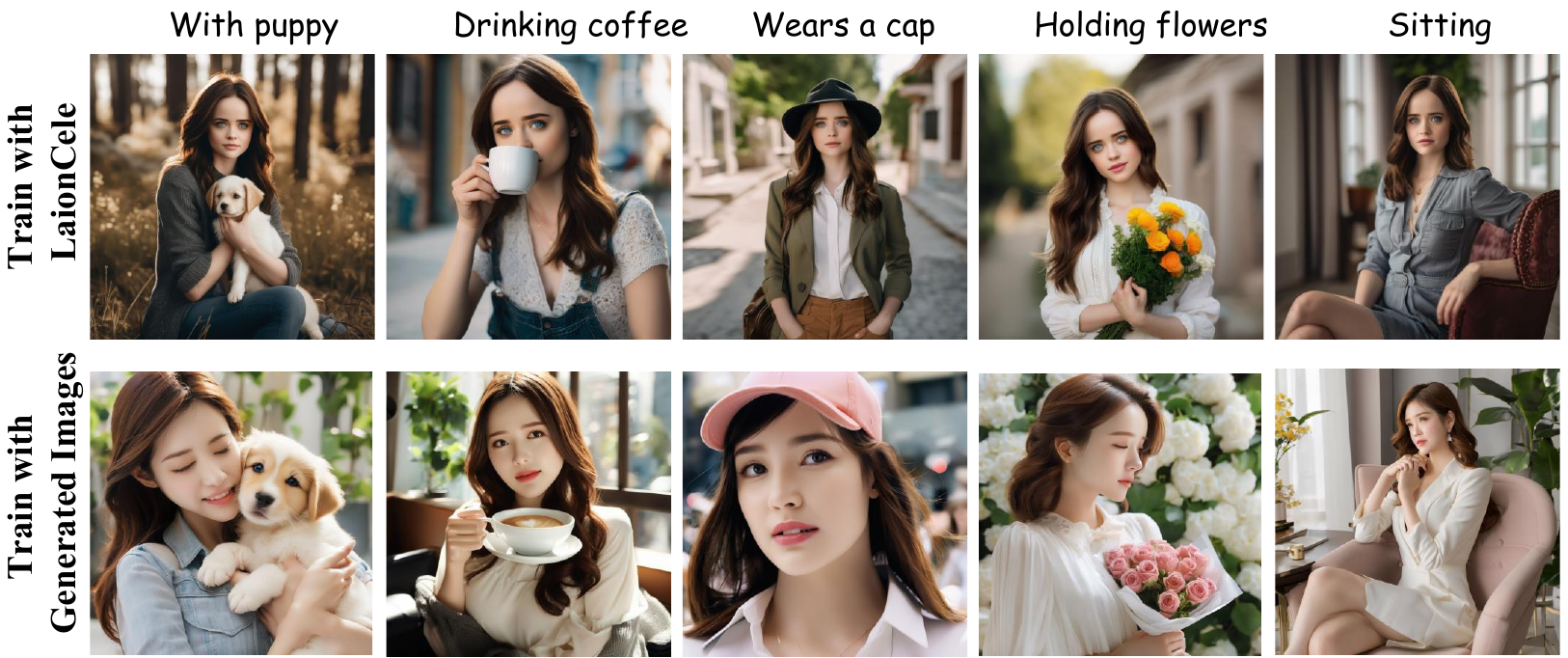}
  \caption{
  The comparison of training an image encoder using the proposed LaionCele dataset and the generated dataset. 
  It is evident that LaionCele is indispensable, and holds substantial value and contribution to the field.
  }
  \label{fig:sdxl_encoder}
\end{figure}

\subsubsection{\textbf{Quantitative Evaluation}}

In our research, ensuring the preservation of individual identity in synthetically generated human images is a fundamental requirement. We concentrate more on maintaining complex semantic consistency in images produced by different generative models, which are driven by challenging textual prompts. Additionally, we place a significant emphasis on the visual appeal of the generated images, striving to achieve a high level of aesthetic quality.

In this study, we performed grouped evaluations of the aforementioned four specified tasks utilizing a pair of evaluation metrics: the CLIP Text-Image Consistency (CLIP-TI) score and the Fréchet Inception Distance (FID). 
Given the widespread acclaim of industry and user for the image generation results of SDXL~\cite{stable_diffusion}, 
we evaluated the similarity of each model's outputs to those of SDXL by calculating their FID score. A lower FID score denotes a higher fidelity to SDXL's synthesis quality and visual appeal. 

Table.~\ref{tab:main1} presents the quantitative results of our evaluation. The outcomes indicate that our approach outperforms other works across all tasks(style, Scene, Emotion, Action) and overall(All) in terms of the CLIP-TI metric. Two points merit particular attention. 
Firstly, our method achieves higher CLIP-TI scores than SDXL due to the latter's inability to recognize name-associated identity attributes, resulting in the omission of character within the scene. The introduction of name embedding in our technique reinforces the concept of person, thereby mitigating the issue of character loss and enhancing semantic consistency.  Secondly, although PhotoMaker's CLIP-TI scores are comparable to those of SDXL, qualitative assessments reveal that PhotoMaker's close-up head images are noticeably inferior in semantic consistency to the well-composed scenes of SDXL.  This suggests that CLIP-TI scores may not effectively discriminate between models in terms of scene construction and visual aesthetic appeal.

Superior similarity with SDXL by FID score indicates exceptional capabilities in scene construction, stylization, emotional editing, action control, and the visual aesthetics of the imagery. The results presented in Table.~\ref{tab:main1}  indicate that our method exhibits a significant advantage in terms of the FID metric. The underlying reason is that our approach does not alter the generator (SDXL) in any way, and the ``name'' prepending operation during the inference process does not induce a shift in the text embedding space. Consequently, this maintains the consistency of identity generation while perfectly preserving the original performance of the generative model.

Table.~\ref{tab:main1} also presents the quantitative assessment results for generated identity consistency. To mitigate the visual bias introduced by effective stylization, as well as the impact of reduced facial features due to the diminution of faces in full-figure presentations within scene construction and action control tasks, we employed the emotional editing task that showcases facial details for the evaluation of identity consistency. The results indicate that our approach also outperforms other methods in terms of generating consistent identities.

\subsection{Create Fictional Identities}
In the  $\mathcal N$  space, any given point corresponds to a person identity. For real individuals, we can obtain their mapping in the $\mathcal N$ space using the image encoder proposed and trained in this paper, thereby achieving consistent identity generation. Furthermore, by interpolating between any two points (i.e., name embedding) within the name space, we can create new fictional characters. As illustrated in Figure~\ref{fig:chazhi}, the appearance of the fictional characters can be flexibly controlled by adjusting the interpolation ratio. Notably, these fictional characters also support consistent identity generation. This will provide significant practical value for film and game production.

\subsection{Ablation Study}

\begin{table}[t]
\caption{Quantitative Evaluation of Fine-Tuning.
}
\label{tabel:fint-tune}
\begin{tabular}{@{}lcccc@{}}
\toprule
                 &    & CLIP-TI ↑ & Ref Simi.↑ & ID Cons.↑ \\ \midrule
\multicolumn{2}{l}{Ours(w/o fine-tune)} & 43.45 &	0.4008 &	0.4407     \\ \hline
                           & LoRA & 37.54&	0.5004	&0.6236    \\
\multirow{-2}{*}{LaionCele}  & U-net &32.55&	0.4475&	0.6631     \\\hline
                           &LoRA    & 40.66&	0.4816&	0.5742     \\ 
\multirow{-2}{*}{Ins700} &U-net   & 36.51	&0.4502	&0.6356     \\ 
\toprule
\end{tabular}
\end{table}

\begin{figure}[t]
  \centering
    \includegraphics[width=\linewidth]{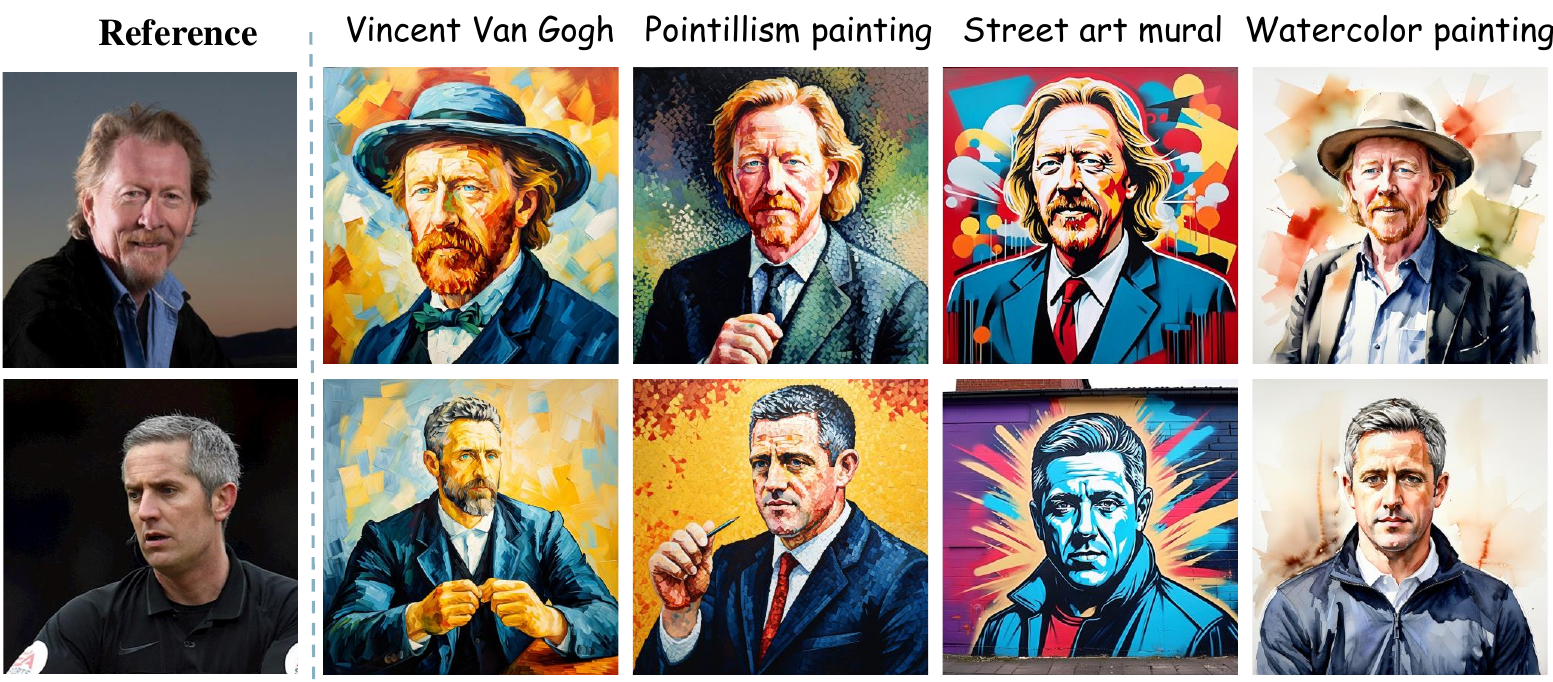}
  \caption{Application on StyleUnet: By integrating a ``name'' into any variant of a stable diffusion, consistent ID generation can be achieved.
  }
  \label{fig:application}
\end{figure}

\subsubsection{\textbf{The necessity of constructing LaionCele datasets.}}
An intuitive approach to construct dataset is leveraging the prior knowledge inherent in SDXL for generation. However, we eschewed this method in favor of an arduous process of filtering and cleaning celebrity image data from the LAION-5B dataset, ultimately yielding a large-scale identity-name dataset, LaionCele.

To assess the disparity between the two datasets (LaionCele and Generated dataset), we also trained an image encoder, $E_{gen}$, with the generated images, and compared its performance with the our image encoder $E_{cele}$ trained with LaionCele. Figure. \ref{fig:sdxl_encoder} illustrates the comparative results. $E_{cele}$ is capable of producing high-quality images with ID and semantic consistency. In contrast, images generated by $E_{gen}$ exhibit numerous artifacts and lack identity consistency.
The results indicate that, on one hand, even well-performing generative models like SDXL exhibit a significant gap between their image outputs and real images. On the other hand, the gap between real test images and generated training images also impacts inference performance. Therefore, it is evident that the dataset constructed in this study, LaionCele, holds substantial value and contribution to the field.

\subsubsection{\textbf{How does fine-tuning impair generative capacity?}}

The primary distinction between our approach and other works lies in our decision not to fine-tune the generative model, thereby preserving its original performance intact. Consequently, one might inquire: how does fine-tuning impair model's  Generative Capacity?

we explored the way of fine-tuning, entire U-net or LoRA on two dataset, i.e., the proposed LaionCele dataset and Ins700(comprising images of 700 individuals scraped from the Instagram website). The scene construction task is employed for evaluation and the results are presented in Table.~\ref{tabel:fint-tune}. The implementation details of fine-tuning are provided in the supplementary material. Experimental results indicate that: 1) Fine-tuning U-net and LoRA both incur a trade-off, enhancing identity consistency(ID Cons.) and reference similarity(Ref Simi.) at the expense of generative performance(CLIP-TI). 2)Fine-tuning U-net results in a more pronounced degradation of generative capability, yet it yields greater improvements in identity consistency. 3) Fine-tuning LoRA is more beneficial for augmenting reference similarity.  4) In comparison to the larger laionCele dataset, smaller datasets(Ins700) exert less influence across all measured metrics.

\subsection{Application: making any U-net identity-aware}
The proposed $\mathcal N$ Space for consistent ID generation is agnostic to the generative model and can be seamlessly integrated with any variant based on SDXL to achieve ID-consistent image generation.
In the aforementioned experiments, our uniformly employed SDXL v1.0 as the generator.
To demonstrate the broad applicability of the $\mathcal N$ Space, we conducted additional experiments on a U-net variant (styleUnet) specialized in style generation.  Figure.~\ref{fig:application} present the consistent ID generation results with styleUnet.
The experimental outcomes indicate that the ``name'' sampled from the $\mathcal N$ Space can be easily utilized with alternative generative models to produce consistent ID generation without compromising the original specialized capabilities. 


\section{Conclusions}
In this paper, we present a novel approach for consistent identity generation of generic identities by introducing a $\mathcal N$ Space. Specifically, we construct a large-scale identity-name dataset, LaionCele, and train an image encoder that maps real images into this $\mathcal N$ Space. This encoder operates independently of the generator, allowing for integration with any SDXL-based generative model to achieve consistent ID generation, offering broad practical applications. A key advantage of our method is that it preserves the original generative capabilities of the generator model, including scene construction, stylization, emotional editing, action control, and so on, thus enhancing the flexibility and creativity of consistent ID generation. Extensive experimental results demonstrated that our approach surpasses existing and concurrent works in terms of ID consistency, semantic consistency, image quality and visual aethetics of the generated images.

\clearpage

\bibliography{aaai25}

\clearpage

\section*{Supplementary Materials}

\subsection{Implementation details}
Our experiments are carried based on  Stableb Diffusion XL v1.0 (stable-diffusion-xl-base-1.0)~\cite{stable_diffusion} with DDIM steps initialized to 100. 
The propose image encoder is trained on a single A800 GPU. The two CLIP encoder in image encoder are clip-vit-large-patch14 and CLIP-ViT-bigG-14-laion2B-39B-b160k-full. 
For training the image encoder, the learning rate was set to 1e-04 with the Adam optimizer.
For the fine-tuning phase, eight A800 GPUs are required, with a LoRA rank set to 16 and a learning rate of 1e-5.
Empirically, the hyper-parameters $\gamma$, $\delta$, and $\eta$ were set to 7, 5, and 80, respectively.

\subsection{Evaluation dataset}
For the quantitative evaluation, we compiled a test dataset consisting of 20 test images and 40 textual prompts. 

For the test images, we did not employ the same test image set as other works~\cite{FastComposer,facestudio,PortraitBooth}, which includes numerous cases of celebrity whose name prompt can guide the generative models, such as SDXL~\cite{stable_diffusion}, to achieve consistent identity generation. 
However, our experiments revealed that the inherent generative capabilities of the generator significantly enhance identity consistency when such cases are used as reference images. Therefore, to genuinely explore the contribution of our work in consistent identity generation, we meticulously selected image cases where SDXL could not generate consistent identities based on name prompts. Since our image encoder training necessitated the use of celebrity cases with consistent identity generation, this also ensured that the test samples were not present in the training set of our image encoder.
The 40 textual prompts is divided into four groups: scene construction, stylization, action control, and emotion editing, with each categroupsgory comprising 10 prompts, as detailed in Table.~\ref{tab1}:

During testing, we invoke the DeepFace~\cite{deepface} model to predict the gender of the input image, subsequently substituting the ``person" in the prompt with ``Man" or ``Woman".
To mitigate the impact of random seed, we generate five images per prompt. Consequently, each test category yields a total of 1,000 synthesized images ($20 \times 10 \times 5$).

\begin{figure}[t]
  \centering
  \includegraphics[width=\linewidth]{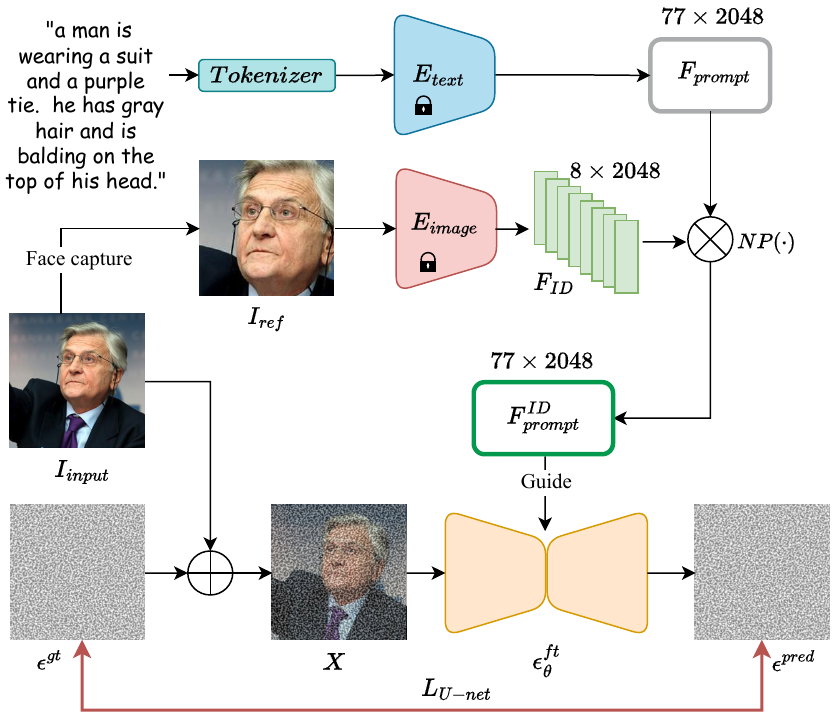}
  \caption{U-Net Fine-Tuning Flowchart. Concurrently with the encoding of prompts via a text encoder, the proposed image encoder is employed to predict the ID features associated with the character. The term $NP(\cdot)$ denotes the operation of feature prepending.
  }
  \label{fig:fine-tune}
\end{figure}

\begin{table*}[t]
\caption{The 40 textual prompts is divided into four groups: scene construction, stylization, action control, and emotion editing, with each categroupsgory comprising 10 prompts.}
\label{tab1}
\begin{tabular}{@{}lll@{}}
\toprule
   & Scene construction                                                           & Stylization                                             \\ \midrule
1  & a person is sitting in a chair                          & a painting of a person in the style of Banksy           \\
2  & a person is   riding a horse                            & a painting of a person in the style of Vincent Van Gogh \\
3  & a person is   drinking coffee                           & a colorful graffiti painting of a person                \\
4  & a person with   a puppy                                 & a watercolor painting of a person                       \\
5  & a person is   skiing                                    & a Greek marble sculpture of a person                    \\
6  & a person is   holding flowers                           & a street art mural of a person                          \\
7  & a person is   holding a camera                          & a black and white photograph of a person                \\
8  & a person is   lying on the lawn                         & a pointillism painting of a person                      \\
9  & a person   leaned on the railing to look at the scenery & a Japanese woodblock print of a person                  \\
10 & a person is   holding an oil-paper umbrella             & a street art stencil of a person                        \\ \midrule
   & Scene construction                                                           & Emotional Editing                                       \\\midrule
1  &       a person in   the jungle                          &  a person is sad                   \\
2  &        a person in the snow                             & a person is angry                                       \\
3  &        a person on the beach                            &        a person is superised       \\
4  &        a person on a cobblestone street                 &     a person is contemptuous       \\
5  &        a person on top of pink fabric                   &         a person is dazed          \\
6  &        a person on top of a wooden floor                &        a person is excited         \\
7  &        a person with a city in the   background         &        a person is suspicious      \\
8  &        a person with a mountain in the   background     &         a person is scared         \\
9  &        a person with a blue house in the   background   &        a person is calm            \\
10 &        a person on top of a purple rug in a   forest    &        a person is happy           \\ \bottomrule
\end{tabular}
\end{table*} 

The CLIP model utilized for computing the CLIP-TI score is CLIP-ViT-bigG-14-laion2B-39B-b160k-full, and identity consistency is computed via the FaceNet~\cite{FaceNet} network.  For a test set comprising 1,000 images, we iteratively process each image and randomly select another image from the remaining 999 to compute the identity consistency. The final identity consistency(ID Cons.) is determined by averaging the values obtained across all 1,000 images.

\begin{algorithm*}[t]
	\caption{MagicNaming: Consistent Identity Generation} 
        \label{algo1}
	\renewcommand{\algorithmicrequire}{\textbf{Input:}}
	\renewcommand{\algorithmicensure}{\textbf{Output:}}
	\begin{algorithmic}[1]
		\REQUIRE A pre-trained Image Encoder $E_{image}$, Text Encoder $E_{text}$ and U-net model $\epsilon_{\theta}$, reference Image $I_{ref}$, ID mean feature $F_{ID}^{mean}$, prompt $p$, hyper-parameters $\gamma$, $\delta$, $\eta$.
		\ENSURE The Generated image $x_0^*$.
            \STATE Get original text embedding $F_{prompt}=E_{text}(T(p))$ .
            \STATE $F_{ID}^{pred} = E_{image}(I_{ref}))$.
            \STATE $F_{ID}^{'} = F_{ID}^{mean} + \delta(F_{ID}^{pred}-F_{ID}^{mean})$.
            \STATE $F_{ID}^{''} = \eta \frac{F_{ID}^{'}}{Norm(F_{ID}^{'}))}$.
            \STATE Integrating name embedding into text embedding, $F_{prompt}^{ID} = NP(F_{ID}^{''},F_{prompt})$.
            \STATE $x_T \sim \mathcal{N}(0,I)$
		\FOR {$t$ from $T$ to $0$ }
            \STATE $\epsilon_t= \epsilon_\theta (x_t | F_{\oslash} ) + \gamma (\epsilon_{\theta}(x_t| F_{prompt}^{ID}) - \epsilon_\theta (x_t | F_{\oslash} ))$
            \STATE $x_{t-1}=\sqrt{\bar{\alpha}_{t-1}}(\frac{x_t-\sqrt{1-\bar{\alpha} _t}\epsilon_t }{\sqrt{\bar{\alpha}_t} } )+\sqrt{1-\bar{\alpha }_{t-1} }\epsilon_t$
		\ENDFOR
		\RETURN $x_0$ as $x_0^*$.
	\end{algorithmic}
\end{algorithm*}

\subsection{Dataset Construction}

The construction of the LaionCele dataset represents one of the significant contributions of this paper, with the detailed construction process summarized in Figure.~\ref{fig:data_process}. The relationship between ``Name" and ``Image" is one-to-many and the operations ``Synchronization'' and ``Filter" in Figure.~\ref{fig:data_process} denote the filtration of multiple images based on a given name or name embedding, and vice versa.  
The critical processing stages encompass the following eight steps:
\begin{enumerate}
    \item Extract name. Utilizing the $en\_core\_web\_sm$ model from spacy, we extracted person names from captions within the Laion5B dataset.
    \item Name cleaning and merging. Utilizing ChatGPT to correct erroneous spellings in names and perform name merging.
    \item Text coding. The text encoder from SDXL is utilized to encode the name, yielding a name embedding. Here, we retain only an 8-dimensional embedding feature for each name.
    \item SDXL generating. Leveraging the ``Name" prepending (NP) method mentioned in the main text, we integrate the name embedding into the embedding features corresponding to the prompt ``a portrait of a person", resulting in the generation of four human portraits.
    \item Evaluating ID consistency. Employing the $face\_recognition$ model, pairwise facial recognition was conducted across four images. Should the images not correspond to the same individual, the respective name embedding were discarded; otherwise, they were retained.
    \item Deduplication. The $average\_hash$ method is employed to deduplicate the authentic images corresponding to the preserved name embedding from the previous step, eliminating the image that are identical in content but vary only in size.
    \item Evaluating ID consistency. Leveraging the $face\_recognition$ model, we discern whether each authentic image corresponds to the generative image associated with the respective name embedding, retaining the authentic image if a match is confirmed, or discarding it otherwise.
    \item Quantity control. To mitigate the long-tail distribution of the dataset, we removed sample pairs of <name embedding-image> where the count of authentic images was less than five.
    
\end{enumerate}

\section{Fine-Tuning Setup}
To investigate how fine-tuning disrupts the performance of generative models, we conducted an ablation study on the fine-tuning way in the main text and presented the validation results. 
Our specific fine-tuning approach is as follows. 

Diverging from the conventional U-Net training methodology, we concurrently encode prompts using a text encoder while capturing the facial region of the input image to feed into our pre-trained image encoder $E_{image}$, enabling it to predict the corresponding name embedding $F_{ID}$ of the subject. Subsequently, we employ  $NP(\cdot)$ operation to obtain $F_{prompt}^{ID}$ infused with the subject's identity information, which then guide the U-Net in predicting noise for the noised image $X$. The network parameters of the U-Net are updated by minimizing the Mean Squared Error (MSE) loss between the actual noise $\epsilon^{gt}$ and the predicted noise $\epsilon^{pred}$, with the loss function defined as follows:
\begin{equation}\label{cfg-unet}
     \mathcal{L}_{U-Net} = \mathbb{E}_{x_t,t,p,\epsilon^{gt}\sim \mathcal{N}(0,1)}||\epsilon_{\theta}(x_t|F_{prompt}^{ID}),\epsilon^{gt}||^2_2
\end{equation}
where $\epsilon_{\theta}(x_t|\cdot)$ is a simplified notation for $\epsilon_{\theta}(x_t,\cdot,t)$, $\epsilon_\theta(\cdot)$ represents the prediction noise of the U-Net model~\cite{Unet15} parameterized by $\theta$. $t$ indicates the time step and $F_{prompt}^{ID}$ is the text embedding with ID information, which is used to guide the generated content. Figure. \ref{fig:fine-tune} illustrates the fine-tuning process.

It is noteworthy that fine-tuning the U-Net is optional. The ``Name'' predicted by the image encoder $E_{image}$ alone can achieve consistent identity generation through $NP(\cdot)$ operation. This paper fine-tuned the U-Net architecture just for the purpose of conducting ablation studies.

\subsection{}section{Algorithm}
we adopt classifier-free guidance in our sampling process, which extrapolates model predictions toward text guidance and away from null-text guidance. 
Specifically, the noise output in the sampling process of classifier-free guidance is computed as follows,
\begin{equation}\label{cfg}
    \epsilon_{\theta}(x_t|F_{prompt}^{ID}) = \epsilon_\theta (x_t | F_{\oslash} ) + \gamma (\epsilon_{\theta}(x_t| F_{prompt}^{ID}) - \epsilon_\theta (x_t | F_{\oslash} ) ),
\end{equation}
where $F_{\oslash}$ represents the null-text embedding and $\gamma$ is the guidance scale used to control the strength of the classifier-free guidance. 

To obtain the final generated image, we follow previous work~\cite{ddpm,ddim} to conduct an iterative denoising process. 
Taking the DDIM~\cite{ddim} sampling for example, a denoising step with $\epsilon_t = \epsilon_{\theta}(x_t|F_{prompt}^{ID}) $ can be denoted as:
\begin{equation}\label{x_t}
    x_{t-1}=\sqrt{\bar{\alpha}_{t-1}}(\frac{x_t-\sqrt{1-\bar{\alpha} _t}\epsilon_t }{\sqrt{\bar{\alpha}_t} } )+\sqrt{1-\bar{\alpha }_{t-1} }\epsilon_t, 
\end{equation}
where $x_t$ is the noisy image in step $t$. $\bar{\alpha}$ is related to a pre-defined variance schedule. 

A concise summary of the consistent identity generation is presented in Algorithm 1.

\subsection{Visualization supplement}

\begin{figure*}[t]
  \centering
  \includegraphics[width=\linewidth]{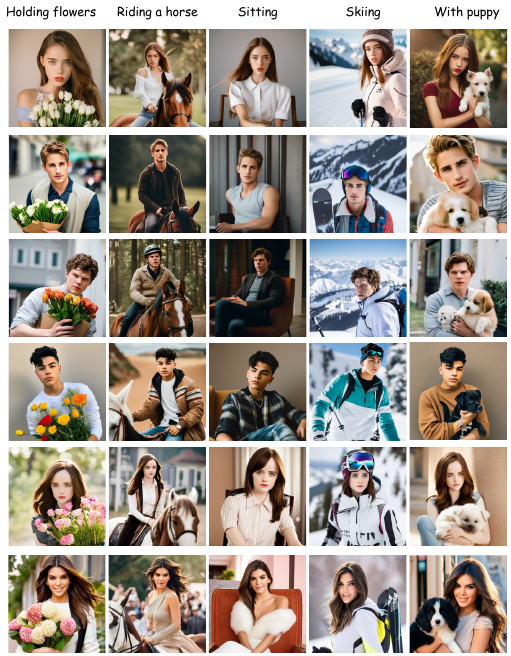}
  \vspace{-1.3em}
  \caption{The experimental data presented in Table 2 of the main text indicate that fine-tuning enhances generative ID consistency. Here, we demonstrate the effects of fine-tuning LoRA on a small dataset web200 dataset using our approach.}
  \label{supfig:lora_web200} 
\end{figure*}

Figure. \ref{subfig:custom} - Figure. \ref{supfig:lora_web200}
present additional visual results. 

\begin{figure*}[t]
  \centering
\includegraphics[width=\linewidth]{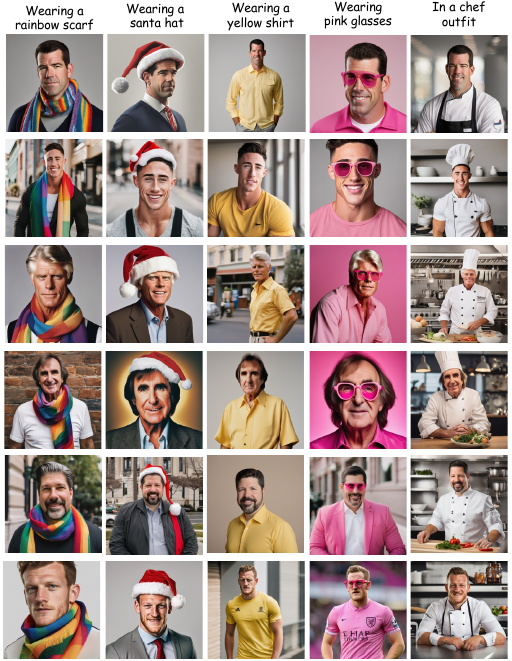}
  \vspace{-1.3em}
  \caption{This figure demonstrates the identity and semantic consistency of our work in the decorative generation tasks.
  }
  \label{supfig:custom}
\end{figure*}

\begin{figure*}[t]
  \centering
  \includegraphics[width=\linewidth]{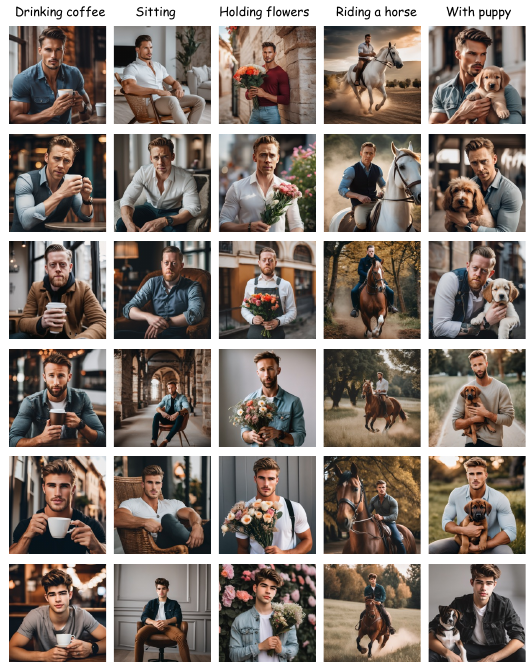}
  \vspace{-2.3em}
  \caption{The supplementary cases demonstrating ID consistency and semantic consistency in the action control generation tasks.
  }
  \label{subfig:action_man} 
\end{figure*}

\begin{figure*}[t]
  \centering
  \includegraphics[width=\linewidth]{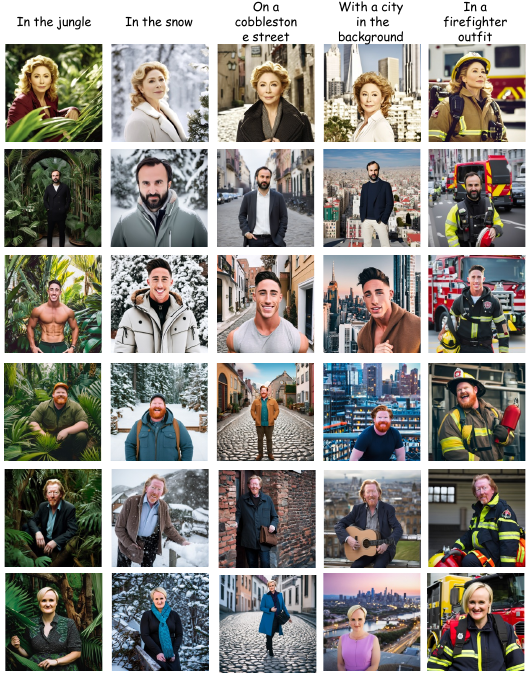}
  \vspace{-1.3em}
  \caption{The experimental data presented in Table 2 of the main text indicate that fine-tuning enhances generative ID consistency. Here, we demonstrate the effects of fine-tuning LoRA on the LaionCele dataset using our approach.
  }
  \label{supfig:loralaion} 
\end{figure*}

\begin{enumerate}
    \item Figure. \ref{supfig:custom} demonstrates the identity and semantic consistency of our work in the decorative generation tasks.
    \item Figure. \ref{subfig:emotion} demonstrates the supplementary cases of emotional editing task.
    \item Figure. \ref{subfig:style_sup} demonstrates the supplementary cases of stylization generation task.
    \item Figure. \ref{subfig:action_man} and Figure. \ref{subfig:action_woman} demonstrates the supplementary cases of action control generation task.
    \item Figure. \ref{supfig:loralaion} demonstrates the ID consistency and semantic consistency of our method with training lora on LaionCele dataset. 
    \item Figure. \ref{supfig:lora_web200} demonstrates the ID consistency and semantic consistency of our method with training lora on a smaller dataset Web200. Fine-tuning on smaller datasets appears to facilitate a more effective balance between identity consistency and image quality.
\end{enumerate}

\begin{figure*}[t]
  \centering
\includegraphics[width=\linewidth]{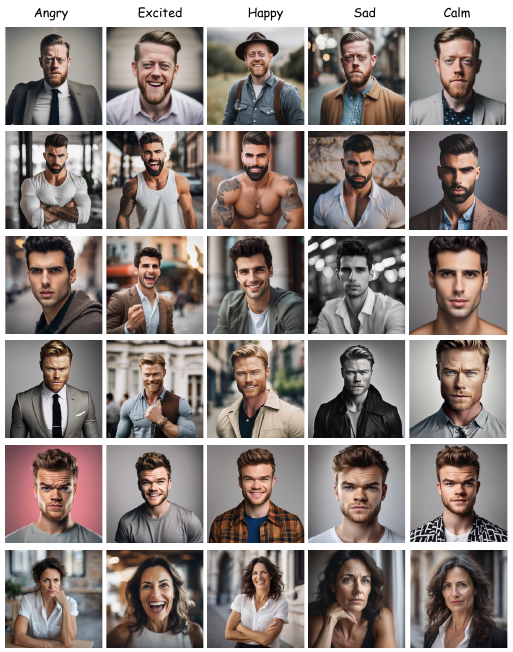}
  \caption{The supplementary cases demonstrating ID consistency and semantic consistency in the emotional editing tasks.
  }
  \label{supfig:emotion}
\end{figure*}

\clearpage

\begin{figure*}[t]
  \centering
  \includegraphics[width=\linewidth]{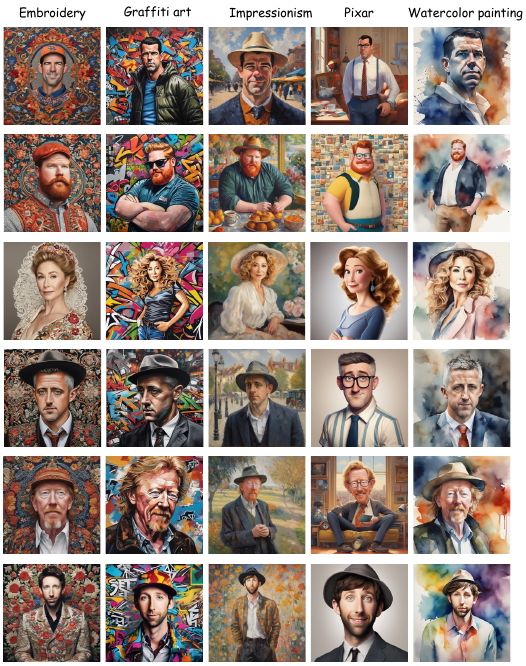}
  \caption{The supplementary cases demonstrating ID consistency and semantic consistency in the stylization generation tasks.}
  \label{supfig:style_sup} 
\end{figure*}
\clearpage

\clearpage
\begin{figure*}[t]
  \centering
  \includegraphics[width=\linewidth]{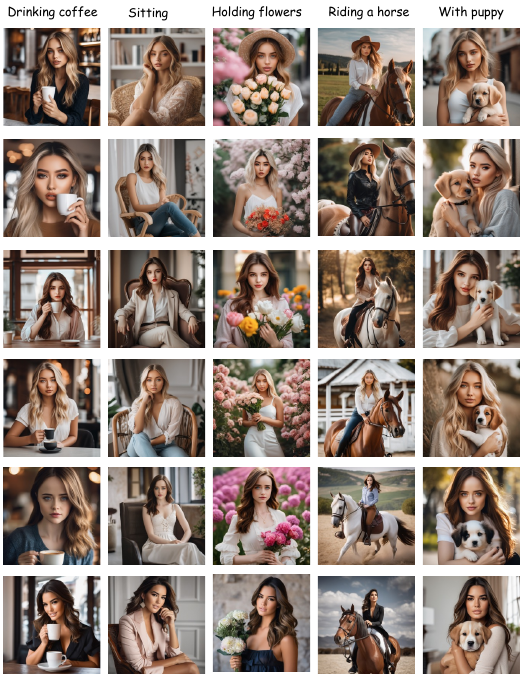}
  \caption{The supplementary cases demonstrating ID consistency and semantic consistency in the action control generation tasks.
  }
  \label{supfig:action_woman} 
\end{figure*}
\clearpage

\clearpage

\begin{figure*}[t]
  \centering
  \includegraphics[width=0.8\linewidth]{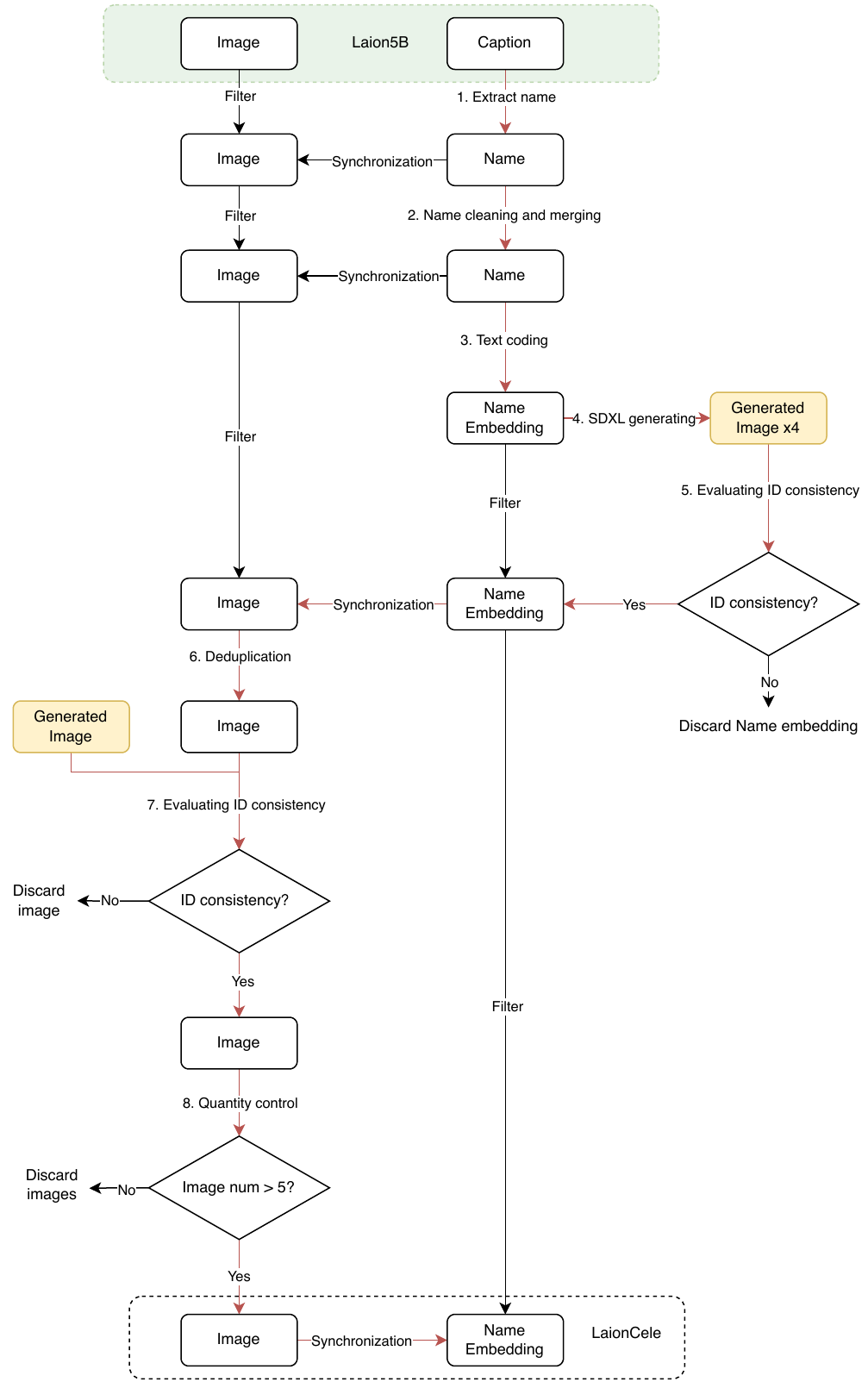}
  \caption{The construction process of LaionCele dataset.
  }
  \label{supfig:data_process}
\end{figure*}

\end{document}